# Analyzing Different Expert-Opined Strategies to Enhance the Effect on the Goal of a Multi-Attribute Decision-Making System Using a Concept of Effort Propagation and Application in Enhancement of High School Students' Performance

Suvojit Dhara
Department of Mathematics, IIT Kharagpur, 721302, West Bengal, India. Tel.: 8240500362
E-mail: sdhara1994@gmail.com

*Adrijit Goswami
Department of Mathematics, IIT Kharagpur, 721302, West Bengal, India. Tel.: 3222-283650
E-mail: goswami@maths.iitkgp.ac.in

**Abstract**

In many real-world multi-attribute decision-making (MADM) problems, mining the inter-relationships and possible hierarchical structures among the factors are considered to be one of the primary tasks. But, besides that, one major task is to determine an optimal strategy to work on the factors to enhance the effect on the goal attribute. This paper proposes two such strategies, namely parallel and hierarchical effort assignment, and propagation strategies. The concept of effort propagation through a strategy is formally defined and described in the paper. Both the parallel and hierarchical strategies are divided into sub-strategies based on whether the assignment of efforts to the factors is uniform or depends upon some appropriate heuristics related to the factors in the system. The adapted and discussed heuristics are the relative significance and effort propagability of the factors. The strategies are analyzed for a real-life case study regarding Indian high school administrative factors that play an important role in enhancing students' performance. Total effort propagation of around 7%-15% to the goal is seen across the proposed strategies given a total of 1 unit of effort to the directly accessible factors of the system. A comparative analysis is adapted to determine the optimal strategy among the proposed ones to enhance student performance most effectively. The highest effort propagation achieved in the work is approximately 14.4348%. The analysis in the paper establishes the necessity of research towards the direction of effort propagation analysis in case of decision-making problems.

## 1. Introduction

In many real-world fields such as business, health sector, educational institutes, etc., decision-making is one of the important tasks to handle. For instance, what are the important factors for better growth of a business, and what should be a proper strategy to enhance the growth? or what are the significant factors that play an important role in improving the performance of a student?

Generally, we refer to the problem of handling these issues as Decision-Making problems. So, structurally, a decision-making problem consists of one or several factors related to a particular goal on which a decision is to be made. In most of the complex problems, we see the number of factors is more than one, in fact in some cases there may be a large number of factors contributing to the decision-making process. Such systems are generally called multi-attribute decision-making (MADM) systems. The primary tasks in case of a MADM system are mining the interrelations among the deciding factors and determination of their system significance, or discovering if there is any hidden hierarchy among the factors or not. These tasks are mainly performed under the supervision of a group of experts chosen based on subject expertise. For example, in a business decision-making process, the experts can be the director, manager, and several other higher rank officers, or for a particular project work, it can be the project supervisor along with several project heads. In health sectors, a particular decision-making process can be supervised by a group of expert doctors, or in the education sector, a group of teachers, professors, or other administrative higher authority can act as the experts. Hence, such decision-making problems can be thought of as an expert-opinion mining task also related to the particular decision to be made. The concerned decision-making methods are called multi-attribute group decision-making (MAGDM) methods [1].

In literature, the most popular MAGDM method to handle the task of mining the inter-relationships among the deciding factors in the system is DEMATEL. The method was first conceptualized by Gabus and Fontela [2] [3]. Then, several of its other upgraded versions such as FDEMATEL [4] [5], IF-DEMATEL [6] [7], etc. did the same task but with more precision. To incorporate the vagueness of the expert opinions, these methods made use of fuzzy [8] or intuitionistic fuzzy [9] linguistics. The methods captured the causal relationships among the factors, i.e., after analyzing the hidden inter-relations among the factors, the factors were classified into two groups "cause" and "effect". In 2023, Dhara and Goswami [10] discussed some structural issues of these methods and came up with a novel decision-making method, called CRRT which not only mined the inter-relationship among the factors, but by finding their system significance, the method ranked the factors as well. To discover a possible hierarchy among the factors, MAGDM methods like ISM [11], AHP [12], ANP [13], etc. were handy. But, Dhara and Goswami showed that in dense systems, i.e., where the factors are densely interrelated, the method ISM completely failed to establish a proper hierarchy among the factors. They also came up with the concept of an iterative hierarchy method, called IHRP [14]. The method was divided into three stages in which the first stage was dedicated to mining the inter-relations among the factors from the expert opinions whereas the second and third stage was there to determine a proper crisp hierarchy among the factors and find their ranking by determining their significance in the system. The method was able to work effectively in the case of dense systems.

We see that in decision-making systems, although the determination of hidden inter-relationships or a probable hierarchy among the factors helps immensely in the decision-making process, but still there remains a vital question to be answered which is 'given the inter-relations or the hierarchy what should be the optimal strategy to act upon the factors to obtain the best results in terms of decision-making?'. That is, 'how much effort is put into the factors and in which pattern' is a critical analysis to make to get the best results for the concerned decision-making process. To the best of our knowledge, the existing research work from this perspective of a decision-making problem is void. In this paper, we propose and analyze two such strategies to enhance the effect on the goal attribute based on a novel concept of effort propagation. Moreover, the strategies were discussed and analyzed for a real-life case study regarding the Indian high school administrative factors which play an important role in the enhancement of students' performance.

The main contribution of the paper is (1) we propose and analyze two different types of effort assignment and propagation strategies, namely parallel and hierarchical strategies, (2) the undertaken case study of analyzing the strategies in the case of Indian high school administrative factors affecting students' performance can be thought of as a novel approach of expert-opinion mining in that direction too. The remainder of the paper is arranged as follows. In section 2, we formally define the problem of effort assignment and propagation. In section 3, the detail of the proposed strategies and their functionality is discussed. Section 4 presents the description of the undertaken case study. The paper concludes with section 5 which discusses the future study aspects of this work as well.

## 2. Problem Definition

This paper deals with the problem of appropriate effort assignment to the factors of a system and propagation through it so that the goal attribute gets maximally affected. We can define "Effort" in terms of the allotment of funds, manpower, or assignment of necessary infrastructures, logistics, etc. In a system, where the factors are inter-connected through hidden relationships/ inter-influences, the allotted effort to a particular factor propagates through various existing paths to other factors, and ultimately some portion of this effort reaches and affects the goal. In decision-making problems, there can be a number of possible assignments of effort to the factors, and they have different impacts on the goal attribute. But in a decision-making environment, our main objective is to enhance the goal maximally. Hence, it becomes quite necessary to determine a proper strategy for this effort assigning, i.e., working on factors so that the goal gets maximally enhanced. In this paper, we propose two expert-opinion-led strategies of assigning efforts to the factors and analyze the effect on the goal attribute using the concept of Effort Propagation and try to find the optimal strategy among them. The formal definition of the problem under consideration is given below.

- ***Problem of finding an optimal strategy to maximize the enhancement of the Goal attribute of the Decision-Making problem***

Given factors $F_1, F_2, \ldots\ldots\ldots F_n$ in a decision-making system, the problem is to find out the optimal strategy of assignment of efforts to factors, that corresponds to the maximum effort propagation to the Goal.

In the next section, we describe various strategies of effort assignment and a novel concept of Effort Propagation mechanism in detail.

## 3. Methodology

For a system where the factors are interconnected through significant influences, there may lie various strategies for effort assignment to the factors which in turn affects the goal in different ways or proportions. In this work, we describe two such strategies and compare these strategies in terms of their effort propagation capabilities. But before analyzing the effort propagation in such strategies, we need to determine the inter-relationships among the factors present in the system and determine the relative significance of the factors in the system. To do that, we can take the help of expert opinions on the factor inter-influences and deploy any of the existing Multi-Attribute Group Decision Making (MAGDM) techniques [15]. We briefly describe the process of mining inter-relationships among the factors in the next subsection.

### 3.1. *Mining Inter-relationship and System Significance of the Factors*

In a system consisting of factors that are interlinked through inter-influences, we can take the help of expert opinions on the factor inter-influences. For that, we need to conduct expert interviews with a questionnaire or approach them with the list of factors and ask them to share opinions on the influence of one factor on another based on some prefixed linguistic scale of influence. If there are $n$ factors in the system, then the opinion data from each expert is in the form of a $n \times n$ matrix. Next, we can deploy any of the MAGDM methods that exist in the literature. The applicable methods for mining the inter-influences of the factors within the system and their system significance values are DEMATEL [16] [17], CRRT [10], ISM [18], IHRP [14], etc. All these methods calculate the weighted aggregation of all the opinion matrices and determine a total relation matrix (TRM). The entry $t(i \to j)$ of this matrix represents the total influence the factor-i has on factor-j. Based on the entries of the TRM, (average of the entries, or sum of the mean and standard deviation of the entries, etc.) a particular threshold value is obtained and the values exceeding the threshold are considered as significant and other values are considered as insignificant. Schemes like DEMATEL, and CRRT dig deeper from this matrix to find several aspects of the inter-factor influences to determine the system significance values of the factors. On the other hand, ISM, IHRP, etc. are hierarchical frameworks that determine the underlying hierarchy among the factors and then determine their system significance values based on their hierarchical positions and inter-influence structures. In the case of dense systems, i.e., systems where the factors are densely inter-influenced, ISM method fails to determine a proper hierarchy [19], but the IHRP method performs well. In the next section, we propose and describe various strategies and discuss the effort propagation in those strategies. The inter-influence and significance values of the factors will be actively used to find the total effort propagation in those strategies.

### 3.2. *Proposed Strategies*

We propose and analyze two different possible strategies as below.

1. Parallel Effort Assignment & Propagation (PEAP) Strategy
2. Hierarchical Effort Assignment & Propagation (HEAP) Strategy

### 3.2.1. *Parallel Effort Assignment & Propagation (PEAP) Strategy*

Among the factors present in a system, some factors are not directly accessible, i.e., we cannot put effort into the factors directly. We can call such factors Latent Factors or Not Directly Accessible Factors (NDAF). Other factors can be directly worked on. Such factors can be referred to as Directly accessible factors (DAF). In the Parallel strategy, we assign efforts to all the directly accessible factors simultaneously and analyze the direct effort propagation to the goal attribute as well as indirect effort propagation to the goal through the latent factors. We assume that effort propagation from one factor to another requires a minimum time gap. Hence, no effort induction/propagation happens between the directly accessible factors as they are worked on simultaneously. *Figure 1* depicts the working mechanism of the PEAP strategy.

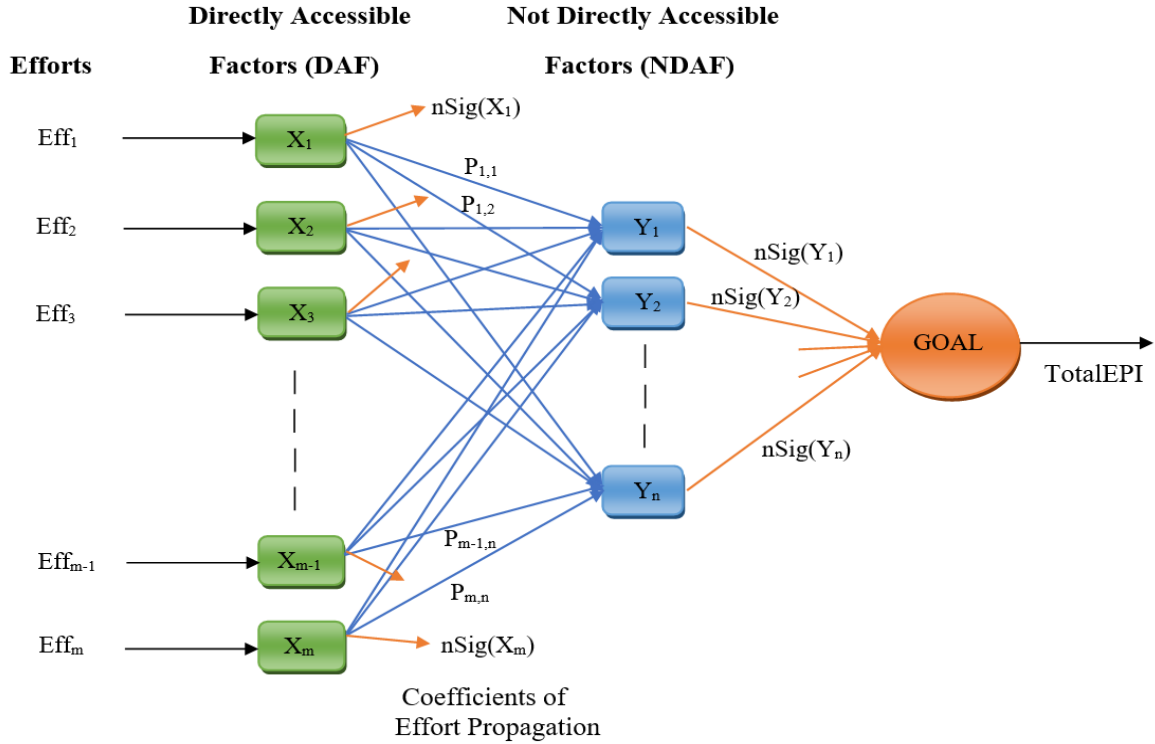

Figure 1. Schematic diagram of the PEAP strategy

The steps for calculation of the total effort propagation in this strategy can be listed as follows.

STEP 1. Assign Efforts to the directly accessible factors. Let, the assigned effort to the factor $F_i$ be $Eff_i$.

STEP 2. If $F_j$ be a latent factor, such that the factor-influence of $F_i$ on $F_j$ is significant, then the propagated effort from $F_i$ to $F_j$ is calculated by the formula given below.

$$EP_{i \rightarrow j} = Eff_i \times d(i \rightarrow j) \qquad (3.1)$$

where $d(i \rightarrow j)$ is the normalized direct influence of $F_i$ on $F_j$.

STEP 3. The effort propagation from the factors to the Goal is directly proportional to their normalized system significance values. Hence, the effort propagated to the goal from the factor $F_i$ can be calculated as below.

$$EP_{i \rightarrow goal} = Eff_i \times nSig(F_i) + \sum_{F_j \in NDAF} EP_{i \rightarrow j} \times nSig(F_j) \qquad (3.2)$$

STEP 4. The total effort propagation index in this strategy is thus calculated as follows.

$$TotalEPI = \sum_{F_i \in DAF} EP_{i \to goal} \qquad (3.3)$$

The normalized direct influence values in equation (3.1) are obtained from the direct influence values between any pair of factors $F_i$ and $F_j$ as follows.

$$d(i \to j) = \frac{D(i \to j)}{\sum_j D(i \to j)} \qquad (3.4)$$

where $D(i \to j)$ is the direct influence value of the factor $F_i$ on $F_j$ obtained after the fuzzy processing of the expert opinion matrices.

As a summary, we can calculate the total effort propagation index ($TotalEPI$) by the following matrix equation

$$TotalEPI = E_{DAF}^T \cdot [nSig_{DAF} + P_{DAF \to NDAF} \cdot nSig_{NDAF}] \qquad (3.5)$$

Here,

$E_{DAF}$ : Actual Effort Matrix of order $m \times 1$

$P_{DAF \to NDAF}$ : Propagation Coefficient Matrix of order $m \times n$

$nSig_{DAF}$ : Normalized Significance matrix (for directly accessible factors) of order $m \times 1$

$nSig_{NDAF}$ : Normalized Significance matrix (for latent factors) of order $n \times 1$

$m$ : Number of directly accessible factors

$n$ : Number of not directly accessible factors

The entries of the $P_{DAF \to NDAF}$ matrix is given as follows

$$P_{ij} = d(i \to j) \qquad (3.6)$$

Two sub-strategies may be derived from here based on the effort assignment to the directly accessible factors. These are

1. Uniform Parallel Effort Assignment & Propagation (U-PEAP)

    The assigned efforts to each of the directly accessible factors are equal and the value of the effort assigned to each of the factors is $\frac{1}{m}$.

2. Weighted Parallel Effort Assignment & Propagation (W-PEAP)

    The assigned efforts to each of the directly accessible factors are directly proportional to their relative system significance values obtained by the hierarchical scheme. For simplicity, we can take the value of the effort assigned to the factor $F_i$ as

$$Eff_i = \frac{Sig(F_i)}{\sum_{F_j \in DAF} Sig(F_j)} \qquad (3.7)$$

In a system where all the factors are directly accessible, only the direct effort propagation from the factors to the goal is considered in the parallel strategy.

### *3.2.2. Hierarchical Effort Assignment & Propagation (HEAP) Strategy*

In hierarchical strategy (HEAP), we deploy the efforts into the factors at different timesteps following a certain hierarchical structure obtained by methods like ISM, IHRP, etc. The total effort remains the same but gets distributed into different timesteps and propagated through different possible paths. The different levels of the underlying hierarchy consisting of one or more factors are defined as different blocks. *Figure 2a*, and *Figure 2b* give an overview of a strategic path in the HEAP strategy.

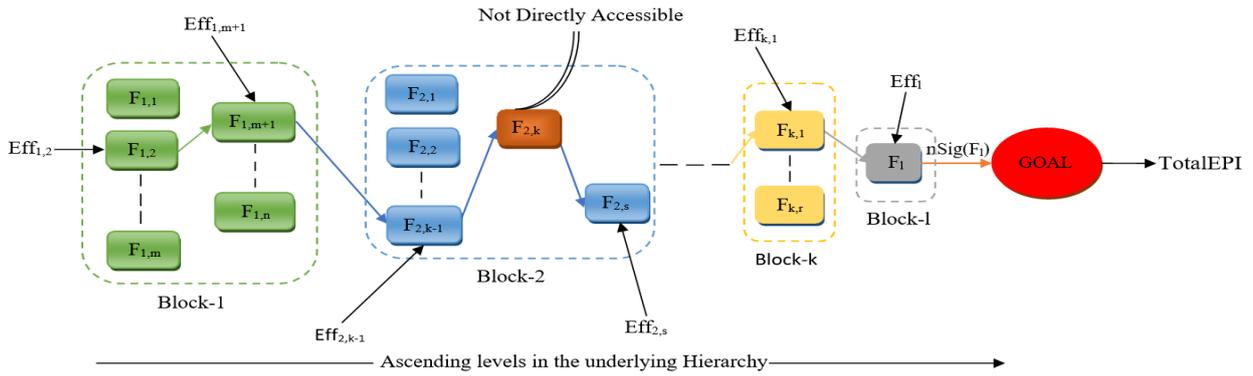

Figure 2a. Overview of a strategic path in HEAP strategy with only one factor from each sublevel

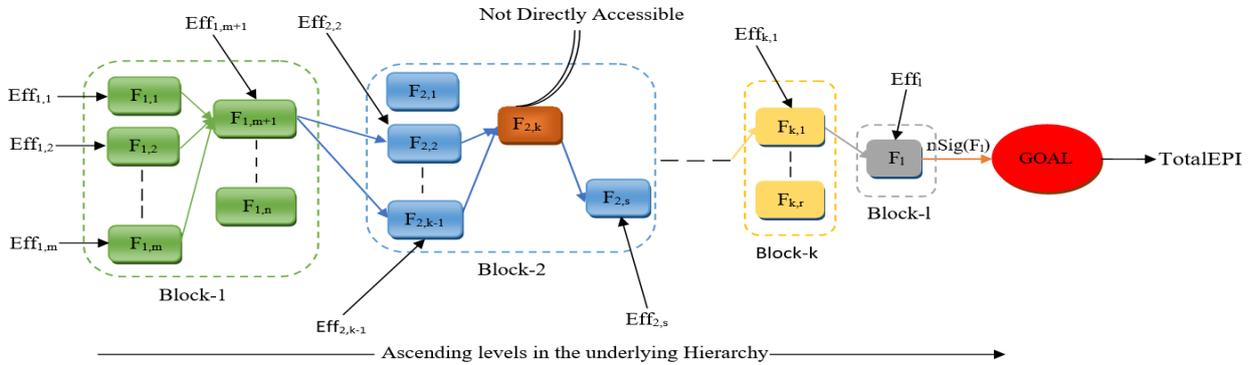

Figure 2b. Overview of a strategic path in HEAP strategy with more than one factor from a sublevel

Before moving on to the proper working procedure of the HEAP strategy, we would like to define a few aspects/basic terms that will be needed for the calculation of the effort propagation in this strategy.

**Definition 3.2.2.1 Direct Effort Propagation (DEP)**

Given a certain amount of effort to a particular directly accessible factor in the strategic path, the amount of effort propagated directly to the goal attribute is known as the direct effort propagation (DEP) of that factor.

For a certain factor, say $F_i$, its direct effort propagation is denoted by $DEP_{F_i}$. This propagation is directly proportional to the normalized system significance of the factor $F_i$. Thus, the value of $DEP_{F_i}$ is calculated by the following formula

$$DEP_{F_i} = Eff_i \times nSig(F_i) \tag{3.8}$$

**Definition 3.2.2.2 Indirect Effort Propagation (IDEP)**

Given a certain amount of effort to a particular directly accessible factor in the strategic path, the amount of effort propagated to the goal attribute of that path through other factors in that path lying in the upper levels (i.e., in the next timesteps) of the underlying hierarchy is known as the indirect effort propagation (IDEP) of that factor.

For a certain factor, say $F_i$, its indirect effort propagation is denoted by $IDEP_{F_i}$. The proportion of the effort assigned to $F_i$ that gets propagated to the goal attribute indirectly through other factors is termed as the indirect effort propagation factor (IDEPF) of $F_i$ and is denoted by $IDEPF_{F_i}$. The value of $IDEP_{F_i}$ is calculated by the following formula

$$IDEP_{F_i} = Eff_i \times \sum_{F_j \in Upper\ level(F_i)} d'(i \to j) \times \{nSig(F_j) + IDEPF_{F_j}\} \tag{3.9}$$

In this formula, $d'(i \to j)$ denotes the influence of $F_i$ on $F_j$ through all possible ascending paths. These values are path-specific and can be determined in a recursive way using the normalized direct influence values among the factors as below.

$$d'(i \to j) = d(i \to j), \quad \text{if } F_i \text{ and } F_j \text{ are in the same or successive blocks}$$

$$d'(i \to j) = d(i \to j) + \sum_{\substack{F_k \text{ is an intermediate} \\ \text{higher-level factor} \\ \text{between } F_i \text{ and } F_j}} d(i \to k) d'(k \to j), \quad \text{else} \tag{3.10}$$

Here, the $IDEPF_{F_i}$ can be recursively computed as

$$IDEPF_{F_l} = 0$$

$$IDEPF_{F_i} = \sum_{F_j \in Upper\ level(F_i)} d'(i \to j) \times \{nSig(F_j) + IDEPF_{F_j}\} \tag{3.11}$$

where, $F_l$ is the last factor in the strategic path.

**Definition 3.2.2.3 Unit Effort Propagation (UEP)**

After assigning a certain amount of effort to a particular directly accessible factor, the amount of effort propagated to the goal attribute is termed as the unit effort propagation (UEP) of that factor.

For a certain factor $F_i$, its unit effort propagation is denoted by $UEP_{F_i}$. This can be calculated by the following equation

$$UEP_{F_i} = Eff_i \times \{nSig(F_i) + IDEPF_{F_i}\} \tag{3.12}$$

The proportion of the effort propagated to the goal attribute from a certain factor can be termed as the unit effort propagation factor (UEPF) of that factor. Then,

$$UEPF_{F_i} = nSig(F_i) + IDEPF_{F_i} \tag{3.13}$$

So, we can write

$$UEP_{F_i} = Eff_i \times UEPF_{F_i} \tag{3.14}$$

The UEPF for the factors can be recursively determined as

$$UEPF_{F_i} = nSig(F_i) + \sum_{F_j \in Upper\ level(F_i)} d'(i \to j) \times UEPF_{F_j} \tag{3.15}$$

**Definition 3.2.2.4 Block Effort Propagation (BEP)**

The amount of effort that gets propagated to the goal attribute from all the factors in a certain block is defined as the block effort propagation (BEP) of that block.

For a certain block $b_i$ of factors, its block effort propagation is denoted by $BEP_{b_i}$. The quantity can be calculated as

$$BEP_{b_i} = \sum_{F_j \in b_i} UEP_{F_j} \tag{3.16}$$

Given the above definitions, now it is easy to arrange the working procedure of the HEAP strategy in the following steps.

STEP 1. Choose a particular strategic path consisting of one or multiple factors from all the sublevels in the underlying hierarchy.

STEP 2. Calculate the number of blocks in the chosen path (say, $p_i$). If the number of blocks is $M$, and if $N$ blocks among them consist of only latent factors, then distribute the total effort to all the $M - N$ effective blocks. Distribute the total effort into these effective blocks.

STEP 3. Calculate the UEP, and BEP values for all the factors and blocks respectively using eq. (3.14), (3.15), and (3.16).

STEP 4. The total effort propagation index for this path $p_i$ in this strategy is calculated as

$$TotalEPI_{p_i} = \sum_{b_j \in p_i} BEP_{b_j} \tag{3.17}$$

As in the case of parallel strategy, here also four sub-strategies may be derived based on the effort assignment to the directly accessible factors at each time step.

1. Uniform block Uniform unit Hierarchical Effort Assignment & Propagation ($U_b U_u$-HEAP)

    The assigned efforts to each of the directly accessible factors in a particular block are equal. If the number of directly accessible factors, which are in the strategic path, in a particular block is k, and the number of effective blocks in the system is $M - N$, then, the value of the effort assigned to each of the factors is $\frac{1}{k(M-N)}$.

2. Uniform block Weighted unit Hierarchical Effort Assignment & Propagation ($U_b W_u$-HEAP)

    The assigned efforts to each of the directly accessible factors in a particular block are directly proportional to their normalized system significance. We can compute the value of the effort assigned to the factor $F_i$ in the strategic path $p$ as

$$Eff_i = \frac{1}{(M-N)} \left[ \frac{nSig(F_i)}{\sum_{F_j \in Block(F_i)\, \cap\, p\, \cap\, DAF} nSig(F_i)} \right] \qquad (3.18)$$

3. Weighted block Uniform unit Hierarchical Effort Assignment & Propagation ($W_b U_u$-HEAP)

    The assigned efforts to each of the directly accessible factors in a particular block are equal to all other factors in the same block. But the total effort assigned to the block is according to the ratio of the sum of the significance values of the factors in the block. Hence, if the number of directly accessible factors in the block of $F_i$, which are part of the path, is k, then, we can compute the value of the effort assigned to the factor $F_i$ as

$$Eff_i = \frac{1}{k} \left[ \frac{\sum_{F_j \in Block(F_i)\, \cap\, p\, \cap\, DAF} nSig(F_j)}{\sum_{F_j \in DAF\, \cap\, p} nSig(F_j)} \right] \qquad (3.19)$$

We can call the expression in the bracket as Block Significance Ratio (BSR) of the block of $F_i$ and can be denoted by $BSR_{Block(F_i)}$.

4. Weighted block Weighted unit Hierarchical Effort Assignment & Propagation ($W_b W_u$-HEAP)

    The assigned efforts to each of the directly accessible factors in a particular block are directly proportional to their UEPF values. Also, we consider the efforts assigned to each block according to its BSR value. Hence, we can compute the value of the effort assigned to the factor $F_i$ as

$$Eff_i = BSR_{Block(F_i)} \times \left[ \frac{nSig(F_i)}{\sum_{F_j \in Block(F_i)\, \cap\, p\, \cap\, DAF} nSig(F_i)} \right] \qquad (3.20)$$

In the hierarchical strategy, we have distributed the efforts into the blocks either uniformly, or based on their BSR value which depends on the sum of the normalized system significance values of the factors in the blocks. Other than this, the efforts can be distributed according to some other heuristics such as the ratio of the average propagability of the blocks (i.e., the average of the UEPF values of all the directly accessible factors in the block). We can call this ratio for a particular block the Block Effort Propagation Ratio (BEPR) of that block. An adjusted combination of both BEPR and BSR can be thought of as another probable heuristic for effort distribution. Similarly, we have considered effort distribution among the factors in a block as either uniform or in the ratio of their normalized system significance values. Effort distribution among the factors in the ratio of their UEPF values can be another heuristic as in that case, we put more effort into the factor that has greater effort propagability. Also, an adjusted combination of both the normalized significance and the UEPF values can be another heuristic. For simplicity, we can represent a hierarchical strategy by the pair of heuristics used for the effort distribution among the blocks and the factors. For instance, our weighted block weighted factor – HEAP strategy can be represented as (BSR, nSig).

## 4. Real-Life Case Study: Application in Case of Indian High School Administrative Factors Affecting Student Performance

Student performance assessment is one of the primary and most important tasks in the field of Educational Data Mining (EDM). If we restrict ourselves to only high school students, then we see besides various student attributes such as his/her demographic data, academic background data, behavioral and non-behavioral data, many of the administrative factors do play some important part in the assessment and improvement of his/her performance [20]. So, A system consisting of some administrative factors which affect student performance significantly can be thought of as a decision-making system where the ultimate decision goal is to enhance student performance by assigning proper efforts to the respective factors. In this section, we discuss the application of the proposed effort propagation strategies in such a decision-making system which consists of some Indian high school administrative factors that affect student performance significantly.

### *4.1. Administrative factors affecting Student Performance*

From the existing research literature on Students' Performance Analysis [21] [22] [23], we see various administrative data such as the number of teachers or students in the school, the professional ability of the teachers, availability of adequate infrastructures, etc. can be thought of as important features when we talk about student performance. Availability of funds is always important in any system and in this case, too it has some salient effect on the improvement of student performance. We see many of the Indian institutes, mainly high schools do not have internet available on the campus. But in foreign countries, we have seen that the use of the Internet while teaching benefits the students immensely. They can use it for their self-study as well. Hence, the availability of the Internet is another feature that we should consider for our decision-making system. There are some features like project and assignment setup and arrangement of doubt-clearing classes for the students which have given impressive results in the case of graduate and post-graduate students in many colleges, and universities across India and the world. So, we can try to incorporate those features in the case of high school students as well and analyze how important a role they play regarding the performance improvement of the students. In *Table 1*, we have enlisted such 18 features [10] which we obtained through extensive research on the literature of student performance assessment.

Table 1. High School Administrative Factors Affecting Student Performance [10]

| Factors | Factors Description |
|---------|---------------------|
| NTeach | No. of teachers in the school |
| NStud | No. of students' intake |
| NSec | No. of sections per class |
| NStaff | No. of non-teaching staff |
| Cln | Cleanliness of the school's environment |
| Pabl | Professional ability of teachers (Classroom teaching, impact on students, classroom resource utilization, etc.) |
| Funds | Funding input |
| Schol | Opportunity for scholarships to students |
| Infs | Teaching infrastructure (Availability of labs, technical instruments, etc.) |
| Assgn | Regular assignment giving and evaluation |
| Prjct | Project setup and evaluation |
| Int | Internet availability on the campus |
| IntTeach | Use of Internet for teaching |
| HighLow | Involving higher-class students in the evaluation of lower class students |
| Doubt | Regular doubt-clearing classes |
| Cocurr | Arrangement for co-curricular activities for students |
| TeachSat | Teachers' satisfaction |
| StudSat | Students' satisfaction |

## 4.2. Data Acquisition and Processing

After the factors in the Indian high school administration that affects student performance were finalized, we approached the experts with the data to seek opinions on the factor inter-influences. The experts were heads of different institutions and some professors associated with teacher training programs in India. The experts took active participation in sharing their opinions. For aggregating the inter-factor influence values, we took the help of the 7-point (0-6) linguistic scale [24] where points represent as follows: 0 – No Influence, 1 – Very low influence, 2 – Low influence, 3 – Medium influence, 4 – High influence, 5 – Very high influence, and 6 – Extremely high influence. All the matrix entries are from this 7-point scale in the opinion data. We collected 20 different expert opinion matrices.

After the data were acquired, we processed the data to mine the inter-influences among the factors and their relative system significance values. As we analyze both the parallel and hierarchical strategies for our system, the IHRP framework [14] was the most appropriate method for the data-mining purpose as it mines the inter-factor influences as well as provides a proper hierarchy among the factors along with their system significance values. To incorporate the vagueness of the expert opinions, we converted the crisp entries into their intuitionistic fuzzy equivalents. We observe that among the factors listed in *Table 1*, the factors "Pabl", "TeachSat" and "StudSat" are not directly accessible and the rest of the factors are directly accessible. The normalized direct influence matrix (N-DIM) among the factors after the fuzzy processing is given in *Table 2*. We have calculated the Total relation matrix (TRM) by the IHRP. As a threshold value, we took the sum of the mean and half of the standard deviation of the entries of TRM, which stands to be 0.1869. The entries in the TRM which exceed this value are considered to be significant.

Table 2. Normalized Direct Influence Matrix (N-DIM)

| Factors | NTeach | NStud | NSec | … | … | TeachSat | StudSat |
|---|---|---|---|---|---|---|---|
| **NTeach** | 0.0109 | 0.0889 | 0.0855 | … | … | 0.0765 | 0.0804 |
| **NStud** | 0.0568 | 0.0086 | 0.0730 | … | … | 0.0627 | 0.0640 |
| **NSec** | 0.0577 | 0.0826 | 0.0096 | … | … | 0.0671 | 0.0674 |
| **…** | … | … | … | … | … | … | … |
| **…** | … | … | … | … | … | … | … |
| **TeachSat** | 0.0520 | 0.0608 | 0.0658 | … | … | 0.0079 | 0.0660 |
| **StudSat** | 0.0590 | 0.0645 | 0.0628 | … | … | 0.0666 | 0.0076 |

After establishing the inter-factor influences using TRM, the IHRP framework digs deeper and finds a proper underlying hierarchical structure among the factors. Then, it determines the relative system significance values of the factors based on their hierarchical positions and their inter-influences. The hierarchy among the factors and their relative system significance values are depicted in *Table 3*.

Table 3. Hierarchical Positions and System Significance of the Factors

| Factors | Hierarchical Levels | Normalized Significance |
|---|---|---|
| **NTeach** | II-A | 0.027303 |
| **NStud** | VII | 0.128171 |
| **NSec** | III | 0.053240 |
| **NStaff** | I-A | 0.004335 |
| **Cln** | I-A | 0.004411 |
| **Pabl** | V | 0.081339 |
| **Funds** | I-C | 0.011092 |
| **Schol** | I-B | 0.008334 |
| **Infs** | VI | 0.101410 |
| **Assgn** | II-B | 0.034508 |
| **Prjct** | IV | 0.065729 |
| **Int** | I-B | 0.008323 |
| **IntTeach** | II-C | 0.043022 |
| **HighLow** | I-A | 0.004411 |
| **Doubt** | I-E | 0.021127 |
| **Cocurr** | I-D | 0.015775 |
| **TeachSat** | VIII | 0.165636 |
| **StudSat** | IX | 0.221834 |

## 4.3. Application of the Effort Assignment & Propagation Strategies

We see among the considered administrative factors, the factors "NStaff", "Cln" and "HighLow" are the least significant ones and they do not have a significant influence on other factors of the system. Thus, in terms of hierarchical positions, they are placed at the lowest sublevel. Hence, we do not consider these factors for effort assignment. The rest 15 factors are considered to be part of the effort assignment and propagation strategy. We discuss and analyze the effort propagation in the proposed strategies within this system below.

### 4.3.1. Analyzing effort propagation in the PEAP strategy

As discussed before, in the PEAP strategy, the directly accessible factors are directly worked on, i.e., put efforts on, and the propagation of this effort to the goal factor via the latent factors is analyzed. In case of our system, among the 15 significant factors, three are not directly accessible and they are "Pabl", "TeachSat", and "StudSat". The remaining 12 factors are directly accessible. *Figure 3* gives a pictorial representation of the PEAP strategy with our high school administrative factors.

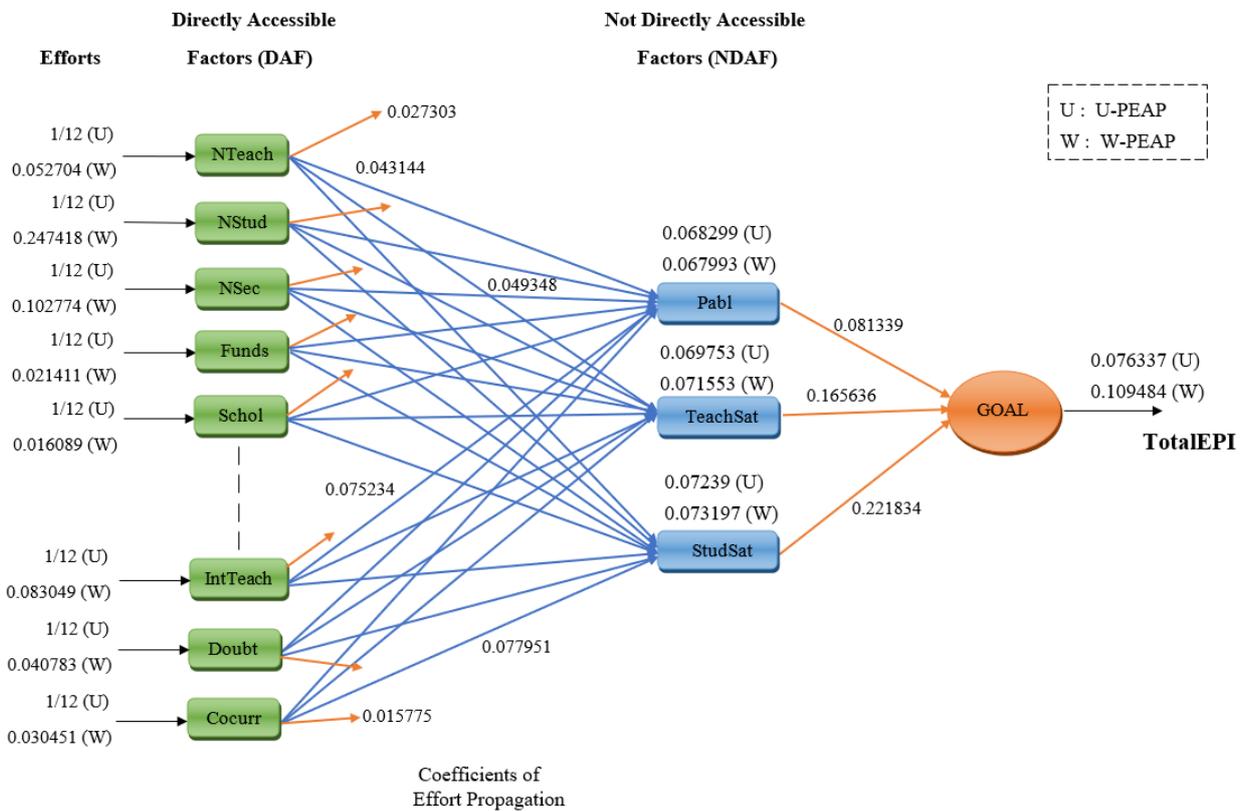

Figure 3. Application of PEAP strategy for the Indian high school administrative factors

*U-PEAP Strategy*

Since there are 12 directly accessible factors in the system which are significant, in the U-PEAP strategy, we apply 1/12 units of effort to each of these factors. We assume the total effort to be 1 unit. The effort propagation coefficients between the directly accessible and the latent factors can be obtained directly from the normalized DIM. Then, we see, the efforts propagated to the latent factors "Pabl", "TeachSat" and "StudSat" are 0.068299 units, 0.069753 units, and 0.07239 units respectively. The significance coefficients for the directly accessible and latent factors can be obtained directly from *Table 2*. Then, the total effort propagation index ($TotalEPI$) for this strategy is 0.076337.

*W-PEAP Strategy*

In the Weighted Parallel (W-PEAP) strategy, the efforts allotted to each of the twelve directly accessible factors are given in *Table 4*. Here also, we assume the total effort to be 1 unit. The effort propagation coefficients and the significance coefficients are the same as before. Then, we see, the efforts propagated to the latent factors "Pabl", "TeachSat" and "StudSat" are 0.067993 units, 0.071553 units, and 0.073197 units respectively. The total effort propagation index ($TotalEPI$) is 0.109484.

Table 4. Effort Assignment to directly accessible factors in W-PEAP strategy

| Factors | Assigned Efforts | Factors | Assigned Efforts |
|---------|------------------|---------|------------------|
| NTeach  | 0.052704         | Assgn   | 0.066613         |
| NStud   | 0.247418         | Prjct   | 0.126881         |
| NSec    | 0.102774         | Int     | 0.016066         |
| Funds   | 0.021411         | IntTeach| 0.083049         |
| Schol   | 0.016089         | Doubt   | 0.040783         |
| Infs    | 0.195760         | Cocurr  | 0.030451         |

*Calculation steps*

To better understand the calculations made in the above strategies, we briefly present a sample calculation of the TotalEPI in the W-PEAP strategy.

Factors affecting the latent factor "Pabl" significantly are "NStud", "NSec", "Infs", "Assgn", "Prjct", "IntTeach", "Doubt", and "Cocurr".

Corresponding effort propagation coefficients are

d(NTeach → Pabl) = 0.043144, d(NStud → Pabl) = 0.061366, d(NSec → Pabl) = 0.049348, d(Funds → Pabl) = 0.051434, d(Schol → Pabl) = 0.076881, d(Infs → Pabl) = 0.076794, d(Assgn → Pabl) = 0.085474, d(Prjct → Pabl) = 0.07412, d(Int → Pabl) = 0.067204, d(IntTeach → Pabl) = 0.075234, d(Doubt → Pabl) = 0.081132, d(Cocurr → Pabl) = 0.077452.

Effort propagated to the factor "Pabl"

= 0.052704×0.043144+0.247418×0.061366+0.102774×0.049348+0.021411×0.051434
+0.016089×0.076881+0.195760×0.076794+0.066613×0.085474+0.126881×0.07412
+0.016066×0.067204+0.083049×0.075234+0.040783×0.081132+0.030451×0.077452

≈ 0.067993

Similarly, we can calculate the efforts propagated to the factors "TeachSat" and "StudSat" and get 0.071553 and 0.073197 respectively.

Then, normalized significance coefficients for the latent factors are

nSig(Pabl) = 0.081339, nSig(TeachSat) = 0.165636, nSig(StudSat) = 0.221834.

The normalized significance coefficients for the directly accessible factors are

nSig(NTeach) = 0.027303, nSig(NStud) = 0.128171, nSig(NSec) = 0.053240, nSig(Funds) = 0.011092, nSig(Schol) = 0.008334, nSig(Infs) = 0.101410, nSig(Assgn) = 0.034508, nSig(Prjct) = 0.065729, nSig(Int) = 0.008323, nSig(IntTeach) = 0.043022, nSig(Doubt) = 0.021127, nSig(Cocurr) = 0.015775

Then, TotalEPI for this strategy

= [0.0527047×0.027303 + 0.247418×0.128171 + 0.102774×0.053240 + 0.021411×0.011092 + 0.016089×0.008334 + 0.195760×0.101410 + 0.066613×0.034508 + 0.126881×0.065729 + 0.016066×0.008323 + 0.083049×0.043022 + 0.040783×0.021127 + 0.030451×0.015775] + [0.067993×0.081339 + 0.071553×0.165636 + 0.073197×0.221834]

≈ 0.109484

### 4.3.2. Analyzing effort propagation in the HEAP strategy

In the hierarchical effort assignment & propagation (HEAP) strategy, the underlying hierarchy is obtained through the IHRP framework. As in the case of the PEAP strategy, here also, we drop the three insignificant factors "NStaff", "Cln" and "HighLow" from the assessment. We see, there are a total of 6 effective blocks. In the first block only, two factors "Schol" and "Int" are placed in the same sublevel and all other factors belong to different sublevels than others. Hence, we see, we can form three strategic paths out of these factors. *Figure 4a*, *Figure 4b*, and *Figure 4c* give an overview of these three strategic paths. Next, we analyze the different HEAP strategies for these paths.

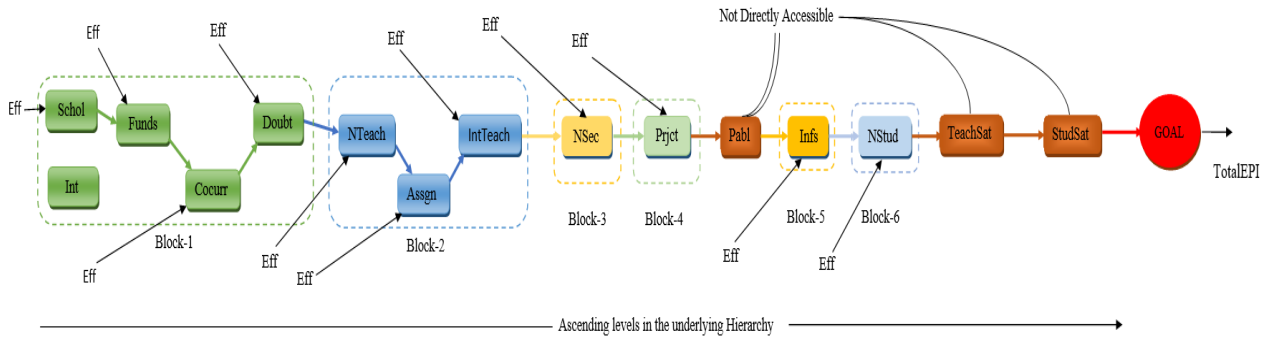

Figure 4a. First Strategic path in the HEAP strategy

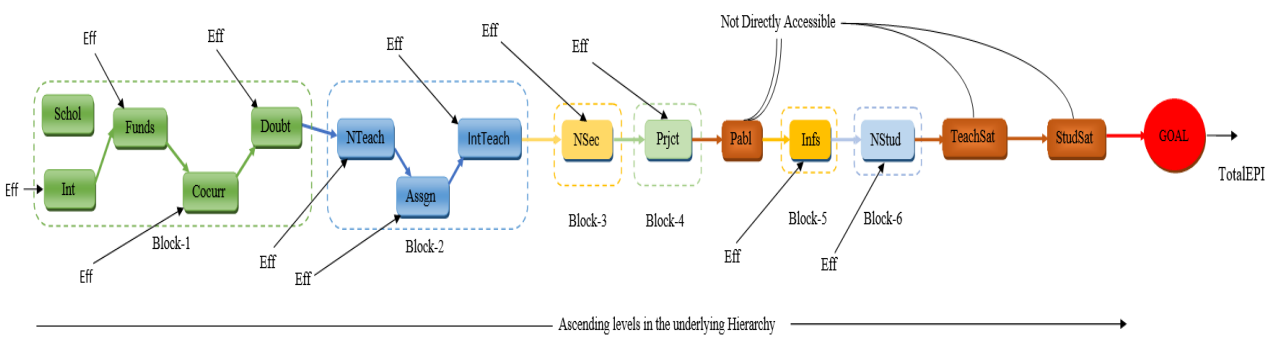

Figure 4b. Second Strategic path in the HEAP strategy

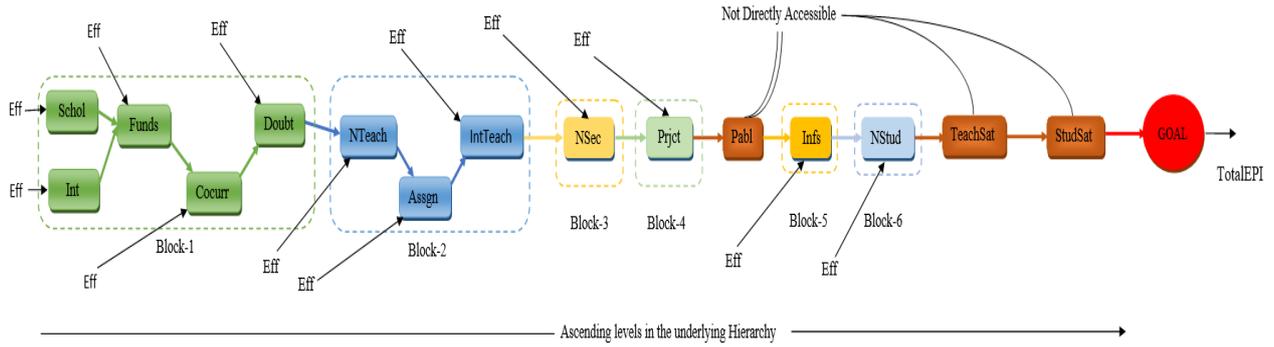

Figure 4c. Third Strategic path in the HEAP strategy

*$U_bU_u$-HEAP Strategy*

In the Uniform block Uniform unit-HEAP strategy, the block coefficient for each of the six blocks is 1/6. In the first and the second path, there are 4 factors in the first block, 3 factors in the second block, and 1 factor in each of the remaining blocks. Hence, efforts assigned to the factors in the first block are 1/24 unit each, that to the factors in the second block are 1/18 unit each, and for the other factors in the remaining blocks are 1/6 unit each. For the third path, there are 5 factors in the first block, and the rest remains the same as in the other two paths. Hence, efforts assigned to the factors in the first block are 1/30 unit each, that to the factors in the second block are 1/18 unit each, and for the other factors in the remaining blocks are 1/6 unit each. The coefficients of effort propagation can be calculated using eq. (3.15). The total effort propagation indexes (TotalEPI) for the first, second, and third strategic paths are 0.142770, 0.142685, and 0.142565.

*$U_bW_u$-HEAP Strategy*

In the Uniform block Weighted unit-HEAP strategy, the block coefficient for each of the six blocks is 1/6. Hence, the effort assigned to each of the blocks is 1/6 unit. The efforts are assigned to the factors in a block in the ratio of their normalized system significance values. For example, the efforts assigned to the factors "Schol", "Funds", "Cocurr", and "Doubt" in the first block of the first strategic path are 0.024659 units, 0.032820 units, 0.046676 units, and 0.062512 units respectively. The other values of the efforts assigned to the factors in any of the blocks in the three paths can be computed similarly. The total effort propagation indexes (TotalEPI) for the first, second, and third strategic paths in this strategy are 0.143447, 0.143397, and 0.143244.

*$W_bU_u$-HEAP Strategy*

In the Weighted block Uniform unit-HEAP strategy, the assignment of efforts happens to the blocks according to the respective BSR values. For instance, the efforts assigned to the blocks for the first strategic path are 0.110510 units, 0.205672 units, 0.104451 units, 0.128953 units, 0.198956 units, and 0.251458 units respectively. Similarly, the block coefficients for other paths can be calculated. The efforts assigned to each of the blocks then get equally distributed among the factors in that block which are part of the path. The total effort propagation indexes (TotalEPI) for the first, second, and third strategic paths in this strategy are 0.143781, 0.143672, and 0.143729.

*W$_b$W$_u$-HEAP Strategy*

In the Weighted block Weighted unit-HEAP strategy, the assignment of efforts to the blocks remains the same as in the previous approach. The efforts assigned to each of the blocks then get distributed among the factors in that block in the ratio of their respective normalized system significance values. The total effort propagation indexes (TotalEPI) for the first, second, and third strategic paths in this strategy are 0.144250, 0.144217, and 0.144348.

*Calculation steps*

In the analysis of the effort propagation through the strategic paths, the calculation of the UEPF for all the factors in the path becomes necessary. We show below how this calculation is done.

Let us consider the first strategic path given in *Figure 4a*.

We see the UEPF value of a particular factor depends on the UEPF values of the factors in the upper levels/blocks. Hence, we start with the calculation of the UEPF value of the last factor in the path, and then the previous one, and so on until the first factor is reached.

The last factor in the first path is "StudSat". We see, nSig(StudSat) = 0.221834, and IDEPF$_{StudSat}$ = 0. So, UEPF$_{StudSat}$ = 0.221834+0 = 0.221834.

The previous factor is "TeachSat". We see, nSig(TeachSat) = 0.165636, d'(TeachSat → StudSat) = 0.065971. Thus, IDEPF$_{TeachSat}$ = 0.065971×0.221834 ≈ 0.014635. Hence, UEPF$_{TeachSat}$ ≈ 0.165636+0.014635 ≈ 0.180271.

The previous factor is "NStud". We see, nSig(NStud) = 0.128171, d'(NStud → TeachSat) = 0.062691, d'(NStud → StudSat) = 0.068104. Thus, IDEPF$_{NStud}$ = 0.062691×0.180271 +0.068104×0.221834 ≈ 0.026409. Hence, UEPF$_{NStud}$ ≈ 0.128171+0.026409 = 0.154580.

Proceeding the same way, we calculate the UEPF values of the previous factors in the path in a recursive way.

The calculation of the BSR values for the blocks in the path is done as follows.

The first block in the first strategic path consists of the factors "Schol", "Funds", "Cocurr" and "Doubt".

We see,

    nSig(Schol) = 0.008334, nSig(Funds) = 0.011092,
    nSig(Cocurr) = 0.015775, nSig(Doubt) = 0.021127

For the other directly accessible factors in the path,

    nSig(NTeach) = 0.027303, nSig(NStud) = 0.128171, nSig(NSec) = 0.053240,
    nSig(Infs) = 0.101410, nSig(Assgn) = 0.034508, nSig(Prjct) = 0.065729,
    nSig(IntTeach) = 0.043022.

Then, the BSR value for the first block of the first strategic path is

$$= \frac{(0.008334+0.011092+0.015775+0.021127)}{(0.008334+0.011092+0.015775+0.021127+0.027303+0.034508+0.043022+0.053240+0.065729+0.101410+0.128171\}}$$

$$\approx 0.110510.$$

## *4.4. Comparative Analysis*

We have applied two proposed strategies to analyze the effort propagation in the high school administration system consisting of factors affecting students' performance. Moreover, the two strategies were divided into several sub-strategies. When we apply equal/uniform efforts to all the directly accessible factors simultaneously, the total effort propagation per unit effort assignment (i.e., we assumed that the total effort assignment is of 1 unit) is approximately 7.6337%. But, when in the same strategy, we go for the assignment of efforts to the factors based on their normalized system significance, then the total effort propagation increases to almost 10.9484%. Again, when we opt for the hierarchical strategy, we obtain different results across different strategic paths. When we distribute the total effort equally to all the blocks and then the factors are also put on efforts uniformly, then from the previous W-PEAP approach, we see a jump in the total effort propagation by almost 30% in the three strategic paths. The maximum effort propagation index observed in this strategy is approximately 14.277% along the first strategic path. Next, the uniform block weighted unit strategy is adopted and we see across all the paths, the total effort propagation gets further increased by 0.4% from the previous best. Next, when the efforts are assigned to the blocks according to their BSR values, and the factors are put on efforts uniformly, then the TotalEPI gets an improvement of a further 0.2%. Lastly, when the efforts were put into the blocks based on their BSR and further into the factors based on their normalized system significance values (nSig), then a further improvement is observed by 0.3% in terms of the total effort propagation. We see, among the applied strategies, the maximum effort propagation to the goal is achieved in the weighted block weighted unit – HEAP strategy under the third strategic path. The TotalEPI in that path is 14.4348%. The results are summarized in *Table 5*.

Table 5. Total Effort Propagation across various proposed strategies for High school Administration factors

| Strategies | Heuristics | TotalEPI |
|---|---|---|
| Parallel Effort Assignment & Propagation (PEAP) | | |
| U-PEAP | | 0.076337 |
| W-PEAP | | 0.109484 |
| Hierarchical Effort Assignment & Propagation (HEAP) | | |
| Uniform block Uniform unit – HEAP | (Uni, Uni) | |
| 1st Strategic Path | | 0.142770 |
| 2nd Strategic Path | | 0.142685 |
| 3rd Strategic Path | | 0.142565 |
| Uniform block Weighted unit – HEAP | (Uni, nSig) | |
| 1st Strategic Path | | 0.143447 |
| 2nd Strategic Path | | 0.143397 |
| 3rd Strategic Path | | 0.143244 |
| Weighted block Uniform unit – HEAP | (BSR, Uni) | |
| 1st Strategic Path | | 0.143781 |
| 2nd Strategic Path | | 0.143672 |
| 3rd Strategic Path | | 0.143729 |
| Weighted block Weighted unit – HEAP | (BSR, nSig) | |
| 1st Strategic Path | | 0.144250 |
| 2nd Strategic Path | | 0.144217 |
| 3rd Strategic Path | | 0.144348 |

We have tried several other heuristics for the effort distribution among the blocks and factors in the hierarchical strategy. The results of the analysis are shown in *Table 6*. We see, generally distributing the efforts to the blocks and then to the factors in a weighted manner increases the total effort propagation to the goal rather than the uniform distribution. Moreover, this can be observed that in the proposed hierarchical strategies, depending on the heuristics used for the distribution of efforts, different strategic paths produce the best TotalEPI value. For instance, in the weighted block weighted unit – HEAP strategies with the heuristics (BEPR, UEPF) and (BEPR, nSig), the maximum TotalEPI is achieved in the first strategic path whereas in the same strategy with heuristics (BSR, UEPF) and (BSR, nSig), the highest TotalEPI is obtained in the third strategic path. The maximum TotalEPI obtained among the applied strategies is 14.4348% which is through the weighted block weighted unit – HEAP strategy with the heuristics (BSR, nSig) along the third strategic path.

Table 6. Effort Propagation across hierarchical strategies with other heuristics

| Strategies | Heuristics | TotalEPI |
|---|---|---|
| Hierarchical Effort Assignment & Propagation (HEAP) | | |
| Uniform block Weighted unit – HEAP | (Uni, UEPF) | |
| 1$^{st}$ Strategic Path | | 0.142927 |
| 2$^{nd}$ Strategic Path | | 0.142848 |
| 3$^{rd}$ Strategic Path | | 0.142702 |
| Weighted block Uniform unit – HEAP | (BEPR, Uni) | |
| 1$^{st}$ Strategic Path | | 0.143530 |
| 2$^{nd}$ Strategic Path | | 0.143430 |
| 3$^{rd}$ Strategic Path | | 0.143288 |
| Weighted block Weighted unit – HEAP | (BSR, UEPF) | |
| 1$^{st}$ Strategic Path | | 0.143848 |
| 2$^{nd}$ Strategic Path | | 0.143795 |
| 3$^{rd}$ Strategic Path | | 0.143896 |
| Weighted block Weighted unit – HEAP | (BEPR, nSig) | |
| 1$^{st}$ Strategic Path | | 0.144251 |
| 2$^{nd}$ Strategic Path | | 0.144197 |
| 3$^{rd}$ Strategic Path | | 0.144028 |
| Weighted block Weighted unit – HEAP | (BEPR, UEPF) | |
| 1$^{st}$ Strategic Path | | 0.143698 |
| 2$^{nd}$ Strategic Path | | 0.143603 |
| 3$^{rd}$ Strategic Path | | 0.143434 |

Overall, it is seen that effort propagation is better when the factors are put on efforts according to the underlying hierarchy than when simultaneously. This happens because putting efforts into the factors step by step allows the efforts to propagate to a greater number of factors in the system.

## 5. Conclusion and Future Works

In real-life decision-making systems which consist of several inter-connected factors, the assessment of the hidden inter-influence or inter-dependencies is an important task. In this regard, if we can determine the underlying hierarchy among the factors, then that helps us to identify the role of each factor in the improvement of the goal. But, after that, the most important task is to analyze and decide how we can work on the factors or how we can assign efforts to the factors to achieve the optimal improvement of the decision. In this paper, we proposed two such effort assignment strategies and analyzed the working pattern of the strategies with the help of the proposed concept of effort propagation. The proposed strategies were validated through a real-life case study in the case of Indian high school administrative factors that affect students' performance. We applied and analyzed the parallel and hierarchical effort assignment and propagation strategies and see that the total effort propagation is highest when we distribute the efforts in several blocks of factors based on their BSR values and put efforts into the factors in the blocks in the ratio of their normalized system significance values. The highest achieved effort propagation index in the analyzed strategies is 14.4348%. Hence, it becomes quite clear that choosing a correct strategy and a correct strategic path is always important to gain better results and achieve better improvement of the goal in a decision-making system.

This work certainly proves the necessity of studying several strategies of effort assignment and propagation in decision-making systems. But the strategies proposed and analyzed in this work are not exhaustive and there are various other possible strategies as well. Hence, we cannot comment that the HEAP strategy or even one of its particular sub-strategies produces maximal effort propagation. One of the other probable strategies is to use multi-variable optimization techniques to find the optimal assignment of efforts to the factors to achieve the highest effort propagation to the goal. Studying other possible strategies should be one of the future works of this study. In the case study, the data is subjective to the expert opinions, and hence if the data pattern changes, the results tend to change according to the data as well. Hence, in the case of the high school administration, or in general for any decision-making system, the application and analysis of the effort assignment and propagation strategies from the very scratch is recommended.

**Acknowledgment**

The authors of this paper would like to express their sincere gratitude to all the teachers and professors who have actively participated in the survey conducted for the data collection for this research work. The work is supported by the Indian Institute of Technology, Kharagpur, India.